\documentclass[twocolumn,english,conference,a4paper]{IEEEtran}
\usepackage[latin9]{inputenc}
\usepackage{color}
\usepackage{babel}
\usepackage{verbatim}
\usepackage{amsmath}
\usepackage{amssymb}
\usepackage{graphicx}
\PassOptionsToPackage{normalem}{ulem}
\usepackage{ulem}
\usepackage[unicode=true,
 bookmarks=true,bookmarksnumbered=true,bookmarksopen=true,bookmarksopenlevel=1,
 breaklinks=false,pdfborder={0 0 0},pdfborderstyle={},backref=false,colorlinks=false]
 {hyperref}
\hypersetup{pdftitle={Your Title},
 pdfauthor={Your Name},
 pdfpagelayout=OneColumn, pdfnewwindow=true, pdfstartview=XYZ, plainpages=false}

\makeatletter
\usepackage[caption=false,font=footnotesize]{subfig}

\usepackage{hyperref}
\hypersetup{
    pdfproducer={},  
 pdfcreator={},   
}

\@ifundefined{showcaptionsetup}{}{%
 \PassOptionsToPackage{caption=false}{subfig}}
\usepackage{subfig}
\makeatother

\begin{document}

\title{Model-based Classification and Novelty Detection For Point Pattern
Data}

\author{\IEEEauthorblockN{Ba-Ngu Vo\IEEEauthorrefmark{1}, Quang N. Tran\IEEEauthorrefmark{1},
Dinh Phung\IEEEauthorrefmark{2} and Ba-Tuong Vo\IEEEauthorrefmark{1}}\IEEEauthorblockA{\IEEEauthorrefmark{1}Curtin University, Australia}\IEEEauthorblockA{\IEEEauthorrefmark{2}Deakin University, Australia}}

\maketitle
{\let\thefootnote\relax\footnotetext{Dinh Phung gratefully acknowledges support from the Air Force Office of Scientific Research under award number FA2386-16-1-4138.}}
\begin{abstract}
Point patterns are sets or multi-sets of unordered elements that can
be found in numerous data sources. However, in data analysis tasks
such as classification and novelty detection, appropriate statistical
models for point pattern data have not received much attention. This
paper proposes the modelling of point pattern data via random finite
sets (RFS). In particular, we propose appropriate likelihood functions,
and a maximum likelihood estimator for learning a tractable family
of RFS models. In novelty detection, we propose novel ranking functions
based on RFS models, which substantially improve performance. %
\end{abstract}


\IEEEpeerreviewmaketitle{}

\textbf{\uline{}}

\section{Introduction}

Point patterns are sets or multi-sets of unordered points (or elements)
that can be found in numerous data sources. In natural language processing
and information retrieval, the `bag-of-words' representation treats
each document as a collection or set of words \cite{joachims1996probabilistic,mccallum1998comparison_NBtextClassifi}.
In image and scene categorization, the `bag-of-visual-words' representation\textendash the
analogue of the `bag-of-words'\textendash treats each image as a set
of its key patches \cite{csurka2004visual}. In data analysis for
the retail industry as well as web management systems, transaction
records such as market-basket data \cite{guha1999rock,yang2002clope}
and web log data \cite{cadez2000EMclustering_VariableLengthData}
are sets of transaction items. Other examples of point pattern data
could be found in drug discovery \cite{dietterich1997solving_multiple_instance},
protein binding site prediction \cite{minhas2012multiple}.

One simple approach to the classification problem for point patterns
is via the naïve Bayes (NB) classifier, see for example \cite{mccallum1998comparison_NBtextClassifi},
\cite{csurka2004visual}, \cite{cadez2000EMclustering_VariableLengthData}.
However, the broader task of learning from point pattern data is more
appropriately posed as a Multiple Instance Learning (MIL) problem
\cite{amores2013multiple_intance_review,foulds2010multi_instance_review},
since multiple instance data or `bags' are indeed point patterns.
According to the recent review article \cite{amores2013multiple_intance_review},
there are three paradigms for multiple instance classification, namely
Instance-Space (IS), Embedded-Space (EM), and Bag-Space (BS). These
paradigms differ in the way they exploit data at the local level (individual
data points within each bag) or at the global level (the bags themselves
as data points). IS is the only paradigm exploiting data at the local
level. At the global level, the ES paradigm maps all point patterns
to vectors of fixed dimension, which are then processed by standard
classifiers for vectors. On the other hand, the BS paradigm addresses
the problem at the most fundamental level by operating directly on
the point patterns. The philosophy of the BS paradigm is to preserve
the information content of the data, which could otherwise be corrupted
through the data transformation process. However, existing methods
in the BS paradigm are confined to distance-based approaches \cite{amores2013multiple_intance_review},
while statistical modelling tools and model-based approaches have
been overlooked. 

In this paper, we introduce statistical models for point pattern data
using Random Finite Set (RFS) theory \cite{Daley88,Moller03statistical,mahler2014advances}.
In particular, we propose appropriate likelihood functions, and a
maximum likelihood estimator for learning (from training data) a tractable
family of models, called iid-cluster RFSs. Further, in novelty detection
where observations are ranked according to their likelihoods, we show
that the standard RFS densities are not suitable for point patterns
and proposed novel ranking functions that substantially improve performance.

\section{A motivating example\label{sec:NB_model} }

\global\long\def\bx{X}

The objective of a classifier is to assign a class label $\hat{y}\in\left\{ 1,\ldots,C\right\} $
to an unseen data point $\bx$ that consists of $m$ feature vectors
$x_{1},...,x_{m}$. In the Bayesian framework, the optimal class label
is given by $\hat{y}=\text{argmax}_{y}p\left(y\mid\bx\right)$, where
$p\left(y\mid\bx\right)\propto p(y)\,p(\bx\mid y)$ is the class posterior
probability, $p(y)$ is the prior probability of class $y$, and\vspace{-1mm}
\begin{align}
p(\bx\mid y) & =p(x_{1},...,x_{m}\mid y)\label{eq:full-joint-likeli}
\end{align}
is the data model or \textit{data likelihood.}

The data model used by the naïve Bayes (NB) classifier \cite[pp. 718]{russell2003artificial}
imposes a conditional independence assumption among the features so
that \vspace{-4mm}

\begin{align}
p(\bx\mid y) & =\prod_{i=1}^{m}p_{f}(x_{i}\mid y)\label{eq:NB_likelihood}
\end{align}
where $p_{f}(x_{i}\mid y)$ is the conditional probability of the
$i$-th feature given class $y$ (see also \cite[pp. 82--89]{murphy2012machine},
\cite[pp. 380--381]{bishop2006pattern}).

In novelty detection or semi-supervised anomaly detection \cite{hodge2004survey},
the data likelihood plays an even more important role. In this approach,
data are ranked according to a data likelihood (learned from normal
data), and data points with likelihoods lower than a threshold are
considered as anomalies \cite{chandola2009anomaly}. 

\begin{figure}[tbh]
\begin{centering}
\vspace{-4mm}
\includegraphics[width=0.9\columnwidth]{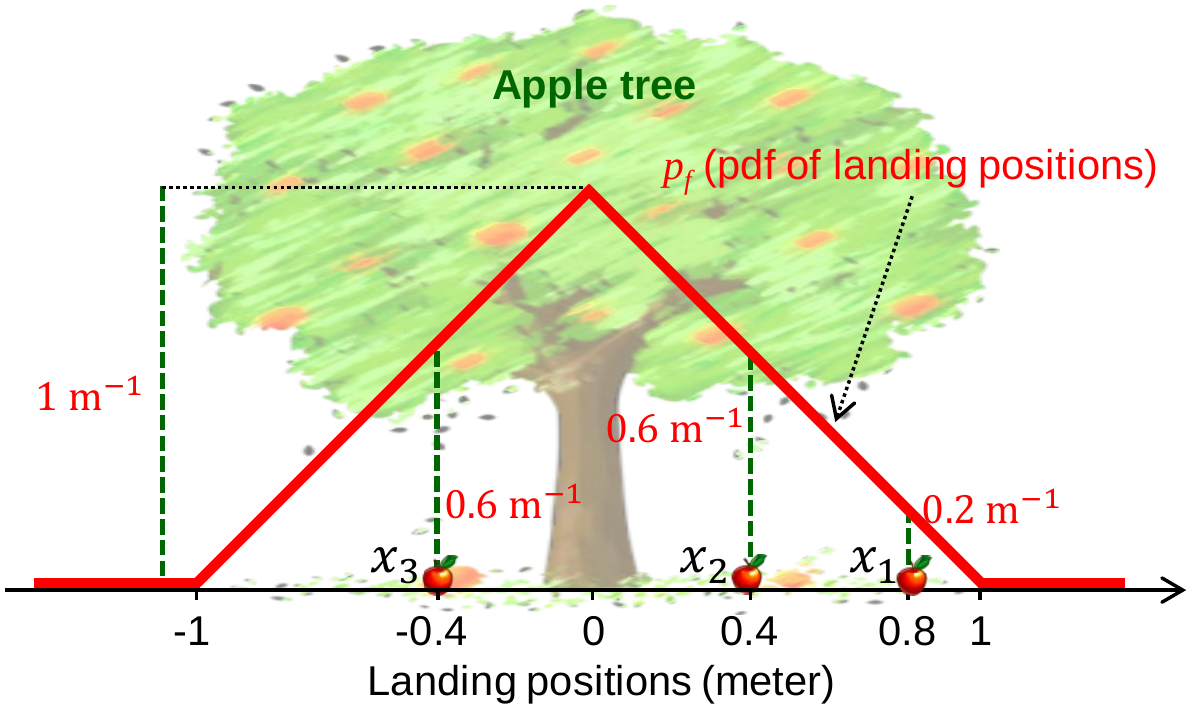}\vspace{-2mm}
\par\end{centering}
\caption{\label{fig:landingPDF}Distribution of landing positions. Position
$x_{1}=0.8\,m$ is 3 times less likely than $x_{2}=0.4\,m$ and $x_{3}=-0.4\,m$
which are equally likely. Credit: clipartbest.com (apple tree clipart)}
\end{figure}

Consider the following example on anomalous patterns of daily fallen
apples. The apples land on the ground independently from each other,
and the probability distribution of the landing positions of the apples
is shown in Fig. \ref{fig:landingPDF} (the thick solid line). The
number of apples and their landing positions for each day are recorded
and we are interested in detecting anomalous behavior of the daily
apple landing pattern. 

Suppose that on day 1 we observed one apple landing at $x_{1}$, and
on day 2 we observed two apples landing at $x_{2}$ and $x_{3}$,
which of the patterns on these two days is more likely to be an anomaly?
To detect anomalies in this setting, it is natural to rank the observed
patterns in order of their likelihoods. In this illustration, we follow
\cite{mccallum1998comparison_NBtextClassifi}, \cite{csurka2004visual},
\cite{cadez2000EMclustering_VariableLengthData} and use the NB likelihood\footnote{For compactness, the condition on the normal class is omitted, i.e.,
$p\left(\bx\right)$ is used instead of $p\left(\bx\mid y=\mbox{`normal'}\right)$.} (\ref{eq:NB_likelihood}) which gives\vspace{-2mm}
\[
p(x_{1})=p_{f}(x_{1})=0.2,
\]
\[
p(x_{2},x_{3})=p_{f}(x_{2})\,p_{f}(x_{3})=0.36.
\]
where $p_{f}$ is pdf of landing positions shown in Fig. \ref{fig:landingPDF}.

Since $p(x_{1})<p(x_{2},x_{3})$, the pattern observed on day 1 \emph{is
more likely }to be an anomaly than the pattern observed on day 2.
However, if we measure distance in centimeters, then\vspace{-2mm}
\[
p(x_{1})=0.002>p(x_{2},x_{3})=0.000036,
\]
and hence the pattern observed on day 2 \emph{is more likely }to be
an anomaly than that of day 1. The likelihood (\ref{eq:NB_likelihood})
yields contradictory results on the same scenario with different units
of measurement! 

Note that in the above analysis, the units of the likelihoods were
overlooked because $p(x_{1})$ is measured in units of $m^{-1}$ or
$cm^{-1}$ while $p(x_{2},x_{3})$ is measured in units of $m^{-2}$
or $cm^{-2}$. The observed inconsistency arises from the incompatibility
in the unit of measurement in the likelihoods $p(x_{1})$ and $p(x_{2},x_{3})$,
i.e., we are not ``comparing apples with apples''. In general, the
unit of the likelihood (\ref{eq:NB_likelihood}) \emph{depends on
number of features in} $\bx$, i.e. the \textit{cardinality} of $\bx$.
Hence, comparing the likelihood (\ref{eq:NB_likelihood}) of different
observations is not meaningful, unless they have the same number of
features or the features are intrinsically unitless. 

\begin{figure}[tbh]
\centering{}\vspace{-2mm}
\includegraphics[width=0.9\columnwidth]{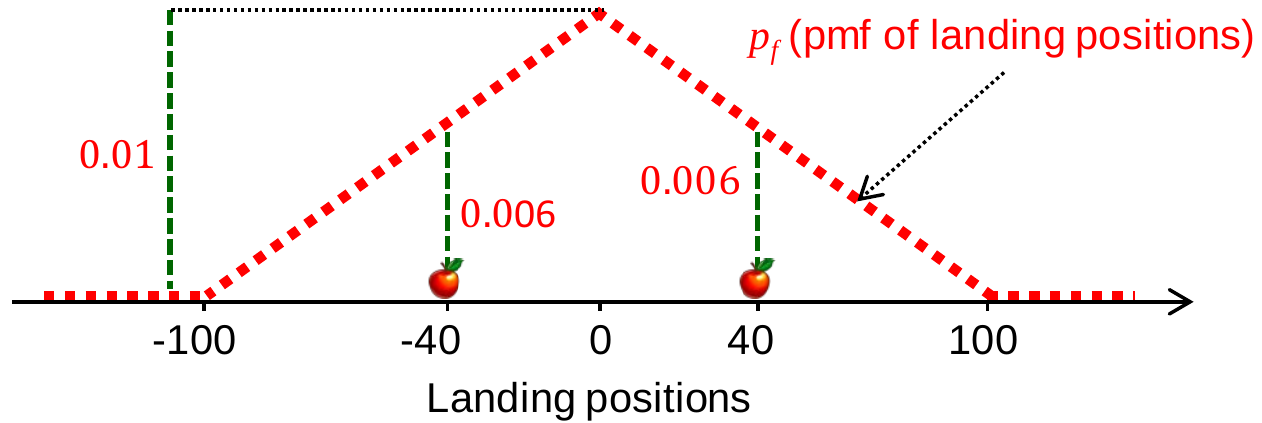}\vspace{-2mm}
\caption{\label{fig:landingPMF_discrete}Distribution of discrete landing positions.}
\end{figure}

Apart from the inconsistency with unit of measurement, such point
pattern likelihood also suffers from another problem\textcolor{black}{{}
associated with cardinality.} Let us revisit the fallen apples example,
however to eliminate the effect of the unit mismatch, this time we
restrict ourselves to a finite number of landing positions, by discretizing
the interval $[-1\,\mathrm{m},1\,\mathrm{m}]$ into 201 points $\left\{ -100,...\,,100\right\} $
and round the landing positions to the nearest of these points (Fig.
\ref{fig:landingPMF_discrete}). Thus, instead of a probability density
of the landing positions on the interval $[-1\,\mathrm{m},1\,\mathrm{m}]$
we now have a (unitless) probability mass function (pmf) on the discrete
set $\left\{ -100,...\,,100\right\} $, see Fig. \ref{fig:landingPMF_discrete}. 

\begin{figure}[tbh]
\begin{centering}
\subfloat[\label{fig:Normal-data}Normal data]{\begin{centering}
\includegraphics[width=0.9\columnwidth]{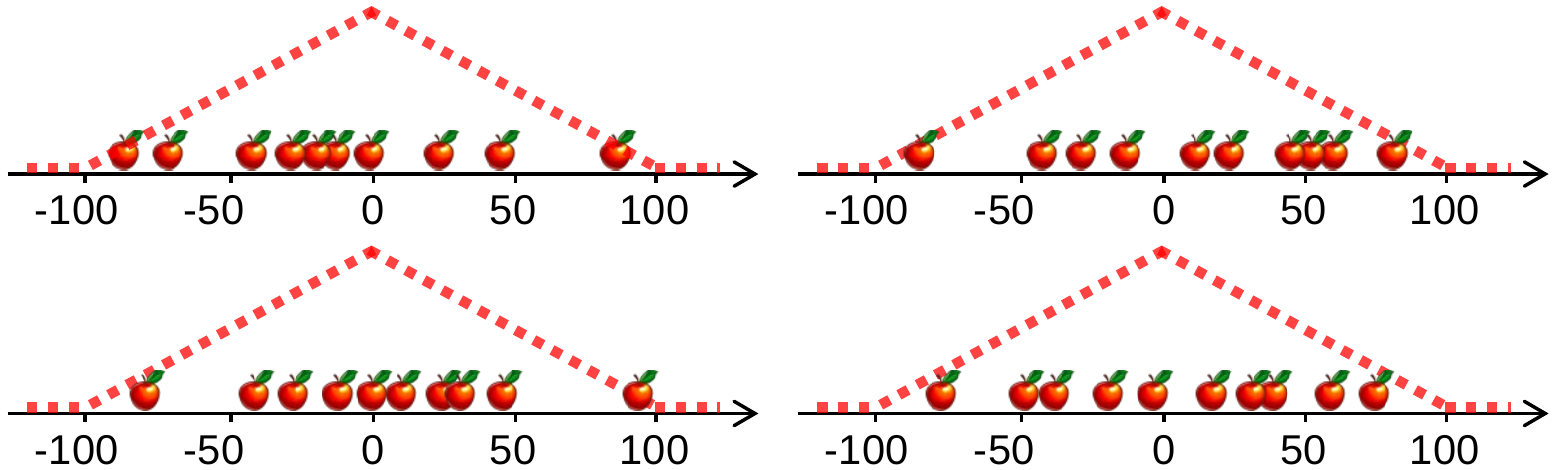}
\par\end{centering}
}\vspace{-2mm}
\subfloat[\label{fig:New-data-points}New observations. Note that, by NB likelihood,
we have $p(X^{(1)})\approx2\times10^{-23}$ and $p(\protect\bx^{(2)})=0.009$]{\begin{centering}
\includegraphics[width=0.9\columnwidth]{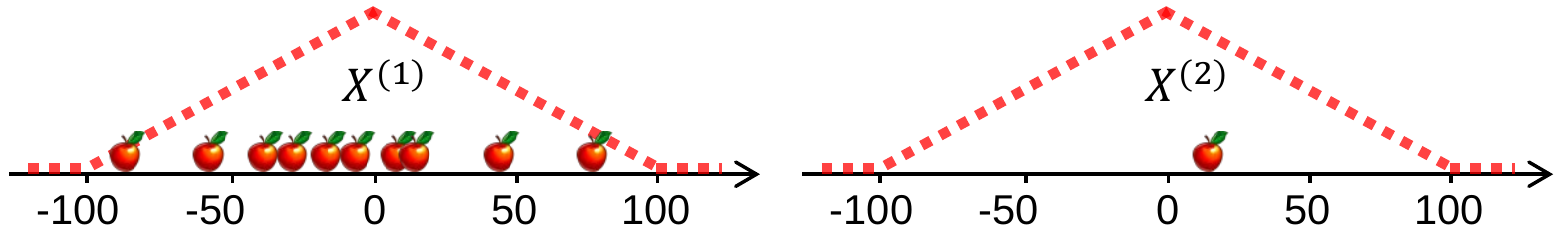}
\par\end{centering}
}
\par\end{centering}
\caption{Examples of normal data and anomaly.}

\vspace{-1mm}
\end{figure}

\textcolor{black}{Fig. \ref{fig:Normal-data} shows 4 `normal' patterns
of fallen apples, each with about 10 locations i.i.d. from the pmf
of Fig. \ref{fig:landingPMF_discrete}. Two new observations $X^{(1)}$
and $X^{(2)}$ whose features are also i.i.d. from the same pmf are
shown in Fig. \ref{fig:New-data-points}.}

Since $X^{(2)}$ has only 1 feature\textcolor{blue}{{} }\textcolor{black}{whereas
the `normal' observations each has around 10 features, it is intuitively
obvious that $X^{(2)}$ is anomalous. However, its likelihood is much
higher than that of $X^{(1)}$ \textendash{} a normal datum }($0.009$
versus $2\times10^{-23}$). This counter intuitive behavior cannot
be attributed to the measurement unit inconsistency because the pmf
of the features is unitless. 

\section{Models for point pattern data }

The likelihood (\ref{eq:NB_likelihood}) was used in the above discussions
to illustrate discrepancies with measurement unit and cardinality.
However, these discrepancies arise even in the full joint likelihood
(\ref{eq:full-joint-likeli}). In this section, we propose models
for point pattern data using random finite set, which could address
these issues. 

\subsection{Random Finite Set\label{subsec:RFS_background}}

Point patterns can be modeled as random finite sets (RFSs), or simple
finite point processes. Point process theory, in general, is concerned
with abstract random counting measures. RFSs are geometrically more
intuitive and thus better suited for the type of discussions in this
article. The likelihood of a point pattern of discrete features is
straightforward since this is simply the product of the cardinality
distribution and the joint probability of the features given the cardinality.
The difficulties arise in continuous feature spaces. In this work,
we only consider continuous feature spaces. 

Let $\mathcal{F}(\mathcal{X})$ denote the space of finite subsets
of a space $\mathcal{X}$. A random finite set (RFS) $X$ of $\mathcal{X}$
is a random variable taking values in $\mathcal{F}(\mathcal{X})$
\cite{Stoyan95,Daley88,Moller03statistical,mahler2007statistical,mahler2014advances}.
In essence, an RFS is a finite-set-valued random variable that is
random in the number of elements, as well as the values of the elements.
An RFS $X$ can be completely specified by a discrete (or categorical)
distribution that characterizes the cardinality $|X|$, and a family
of symmetric joint distributions that characterizes the distribution
of the points (or features) of $X$, conditional on the cardinality.

Analogous to random vectors, the probability density of an RFS (if
it exists) is essential in the modeling of point pattern data. The
probability density $p:\mathcal{F}(\mathcal{X})\rightarrow[0,\infty)$
of an RFS is the Radon-Nikodym derivative of its probability distribution
relative to the dominating measure $\mu$, defined for each (measurable)
$\mathcal{T}$ \ensuremath{\subseteq} $\mathcal{F}(\mathcal{X})$,
by \cite{Stoyan95,Moller03statistical,vo2005sequential,hoang2015cauchy_schwarz}:\vspace{-2mm}
\begin{align}
 & \mu(\mathcal{T})=\nonumber \\
 & \sum_{m=0}^{\infty}\frac{1}{m!U^{m}}\int\mathbf{1}_{\mathcal{T}}\left(\{x_{1},...,x_{m}\}\right)d\left(x_{1},...,x_{m}\right)\label{eq:commonRefMeasure}
\end{align}
where $U$ is the unit of hyper-volume in $\mathcal{X}$, and $\mathbf{1}_{\mathcal{T}}(\cdot)$
is the indicator function for $\mathcal{T}$. The measure $\mu$ is
the unnormalized distribution of a Poisson point process with unit
intensity $u=1/U$ when $\mathcal{X}$ is bounded. Note that $\mu$
is unitless and consequently the probability density $p$ is also
unitless. 

In general the probability density of an RFS, with respect to $\mu$,
evaluated at $X=\{x_{1},...,x_{m}\}$ can be written as
\begin{equation}
p(X)=p_{c}(m)\,m!\,U^{m}f_{m}\left(x_{1},...,x_{m}\right),\label{eq:generalRFSdensity}
\end{equation}
where $p_{c}(m)=\mbox{Pr}(|X|=m)$ is the cardinality distribution,
and $f_{m}\left(x_{1},...,x_{m}\right)$ is a symmetric joint probability
density of the points $x_{1},...,x_{m}$ given the cardinality,\textcolor{black}{{}
see \cite[p. 27]{lieshout2000markov} ((Eqs. (1.5), (1.6), and (1.7)),
\cite{Moller03statistical}, \cite{vo2005sequential}.}

\subsection{Likelihoods for point pattern data}

Instead of a random vector $\bx$, we propose to model each point
pattern as an RFS $X$. A general form for the likelihood of $X$
is given by (\ref{eq:generalRFSdensity}), which can capture the cardinality
information as well as the dependence between the features. The RFS
data model also avoids the unit of measurement inconsistency since
the probability density with respect to $\mu$ is unitless. 

Imposing the `naïve' conditional independence assumption among the
features on the model in (\ref{eq:generalRFSdensity}) reduces to
the \emph{iid-cluster RFS} model \cite{Daley88}\vspace{-1mm}
\begin{equation}
p(X)=p_{c}(|X|)\,|X|!\,[Up_{f}]^{X}\label{eq:iidRFSdensity}
\end{equation}
where $p_{f}$ is a probability density on $\mathcal{X}$, referred
to as the\emph{ feature density, }and $h^{X}\triangleq\prod_{x\in X}h(x)$,
with $h^{\emptyset}=1$ by convention, is the finite-set exponential
notation.

When $p_{c}$ is a Poisson distribution we have the celebrated \emph{Poisson
point process} (aka, \emph{Poisson RFS})\vspace{-1mm}
\begin{equation}
p(X)=\lambda^{|X|}\,e^{-\lambda}\,[Up_{f}]^{X}
\end{equation}
where $\lambda$ is the mean cardinality. The Poisson model is completely
determined by the intensity function $u=\lambda p_{f}$ \cite{Moller03statistical},\cite{vo2005sequential,hoang2015cauchy_schwarz}.
Note that the Poisson cardinality distribution is described by a single
non-negative number $\lambda$, hence there is only one degree of
freedom in the choice of cardinality distribution for the Poisson
model.

Given the training data, a key task in learning is to compute estimates
of the underlying parameters of the model. Learning the general model
(\ref{eq:generalRFSdensity}) is computationally intensive. The iid-cluster
model (\ref{eq:iidRFSdensity}), on the other hand, provides a good
trade-off between tractability and flexibility. 

\subsection{Maximum Likelihood Estimation\label{subsec:Learn_RFS_params}}

This subsection presents a solution to learning the parameters of
an iid-cluster RFS model using maximum likelihood (ML) estimation. 

Given a finite list of observations $Z^{(1)},...,Z^{(N)}\in\mathcal{Z}$
and a parametrized probability density $f(\cdot\negthinspace\mid\theta)$
on $\mathcal{Z}$, we denote\vspace{-2mm}
\begin{equation}
\hat{\theta}(f;Z^{(1)},...,Z^{(N)})\triangleq\underset{\theta}{\mbox{argmax}}\left(\prod_{n=1}^{N}f(Z^{(n)}\mid\theta)\right)\label{eq:MLE_general}
\end{equation}
If the data points $Z^{(1)},...,Z^{(N)}$ are i.i.d. according to
$f(\cdot\negthinspace\mid\theta)$, then $\hat{\theta}(f;Z^{(1)},...,Z^{(N)})$
is indeed the maximum likelihood estimate (MLE) of $\theta$. 

Since an iid-cluster RFS is uniquely determined by its cardinality
and feature distributions, we consider cardinality and feature distributions
parametrized by $p_{c}(\cdot|\theta_{c})$ and $p_{f}(\cdot|\theta_{f})$,
i.e.\vspace{-2mm}
\begin{align}
p(X\mid\theta_{c},\theta_{f})= & p_{c}(|X|\mid\theta_{c})\,|X|!\,U^{|X|}\left[p_{f}(\cdot|\theta_{f})\right]^{X}\label{eq:iidRFSdensity_wParams}
\end{align}
Learning the underlying parameters of an iid-cluster model amounts
to estimating $\theta=(\theta_{c},\theta_{f})$ from training data.
Furthermore, the MLE of the iid-cluster model parameters separates
into the MLE of the cardinality distribution parameters $\theta_{c}$
and MLE of the feature density parameters $\theta_{f}$. This is stated
more concisely in the following Proposition.

\textbf{Proposition 1.} Let $X^{(1)},...,X^{(N)}$ be $N$ i.i.d.
realizations of an iid-cluster RFS with parametrized cardinality distribution
$p_{c}(\cdot|\theta_{c})$ and feature density $p_{f}(\cdot|\theta_{f})$.
Then the MLE of $(\theta_{c},\theta_{f}),$ is given by
\begin{eqnarray}
\hat{\theta}_{c} & = & \hat{\theta}\left(p_{c};|X^{(1)}|,...,|X^{(N)}|\right)\label{eq:Prop1_1}\\
\hat{\theta}_{f} & = & \hat{\theta}\left(p_{f};\uplus_{n=1}^{N}X^{(n)}\right)\label{eq:Prop1_2}
\end{eqnarray}
where $\uplus_{n=1}^{N}X^{(n)}$ is the disjoint union of $X^{(1)},...,X^{(N)}$.

\emph{Proof.} Using (\ref{eq:iidRFSdensity_wParams}), we have\vspace{-2mm}
\begin{align}
\hspace{-6mm} & \prod_{n=1}^{N}p(X^{(n)}\mid\theta_{c},\theta_{f})\nonumber \\
\hspace{-6mm} & =\prod_{n=1}^{N}p_{c}(|X^{(n)}|\mid\theta_{c})\,|X^{(n)}|!\,U^{|X^{(n)}|}\prod_{x\in X^{(n)}}p_{f}(x\mid\theta_{f})\nonumber \\
\hspace{-6mm} & =\left(\prod_{n=1}^{N}p_{c}(|X^{(n)}|\mid\theta_{c})\right)\cdot\left(\prod_{n=1}^{N}|X^{(n)}|!\,U^{|X^{(n)}|}\right)\nonumber \\
\hspace{-6mm} & \,\,\,\cdot\left(\prod_{n=1}^{N}\prod_{x\in X^{(n)}}p_{f}(x\mid\theta_{f})\right)\label{eq:MLE_proof}
\end{align}
Hence, to maximize the likelihood we simply maximize the first and
last bracketed terms in (\ref{eq:MLE_proof}) separately. This is
achieved with (\ref{eq:Prop1_1}) and (\ref{eq:Prop1_2}). QED.

Observed from Proposition 1 that the MLE of the feature density parameters
is identical to that used in NB. \textcolor{black}{For example, if
the feature density is a Gaussian $\mathcal{N}\left(\boldsymbol{\mu},\boldsymbol{\Sigma}\right)$,
then the parameters of its ML estimates are:
\begin{align}
\hat{\boldsymbol{\mu}} & =\frac{1}{N}{\textstyle \sum}_{n=1}^{N}{\textstyle \sum}_{x\in X^{(n)}}x,\\
\hat{\boldsymbol{\Sigma}} & =\frac{1}{N}{\textstyle \sum}_{n=1}^{N}{\textstyle \sum}_{x\in X^{(n)}}\left(x-\hat{\boldsymbol{\mu}}\right)\left(x-\hat{\boldsymbol{\mu}}\right)^{T}.
\end{align}
}Consequently, the iid-cluster model requires only one additional
task of computing the MLE of the cardinality distribution parameters,
which is relatively inexpensive. 

For a categorical cardinality distribution, i.e. $\theta_{c}=\left(p_{1},...,p_{K}\right)$,
the MLE of the cardinality distribution is given by\vspace{-3mm}
\begin{eqnarray}
\hat{p}_{k} & = & \frac{1}{N}{\textstyle \sum}_{n=1}^{N}\delta_{k}(|X^{(n)}|).
\end{eqnarray}
Since there are $K$ parameters $p_{1},...,p_{K}$, we require a sufficiently
large dataset (significantly larger than $K$). For a small dataset,
a cardinality distribution with a small number of parameters should
be used to avoid over-fitting, e.g. Poisson, i.e. $\theta_{c}=\left(\lambda\right)$
where its MLE is given by\vspace{-2mm}
\begin{equation}
\hat{\lambda}=\frac{1}{N}{\textstyle \sum}_{n=1}^{N}|X^{(n)}|.
\end{equation}
Conceptually, Proposition 1 can be extended to the general RFS model
(\ref{eq:generalRFSdensity}), which relaxes the naïve independence
assumption. The MLE of the cardinality distribution parameters is
computed as for the iid-cluster RFS model. However, for the feature
distribution, instead of $\hat{\theta}(p_{f};\uplus_{n=1}^{N}X^{(n)})$
we need to compute the MLE of the parameters for the joint densities,
i.e. \vspace{-2mm}
\begin{equation}
\underset{\theta}{\mbox{argmax}}\prod_{n=1}^{N}f_{|X^{(n)}|}(X^{(n)}\mid\theta_{f})
\end{equation}
which is far more complex in general. Imposing additional assumptions
such as TAN may provide some simplifications. Alternative models such
as mixture of iid-cluster RFSs \cite{PV_RFSMODEL_FUSION14} are also
promising. However, these are topics for future research.

\subsection{Numerical experiments\label{subsec:experiments_RFS}}

This subsection presents two classification experiments with simulated
and real-world data. These experiments involve learning from training
data, and then use the learned models to classify new observations.
The results are bench-marked against the NB model. The first experiment
uses simulated data so as to illustrate the benefit of cardinality
information when the features between the classes are similarly distributed.
The second experiment uses real-world data from the Texture images
dataset \cite{textureDataset}.

\subsubsection{Classification with simulated data\label{subsec:Classification-with-simulated}}

\begin{figure}[tbh]
\begin{centering}
\vspace{-3mm}
\includegraphics[width=0.9\columnwidth]{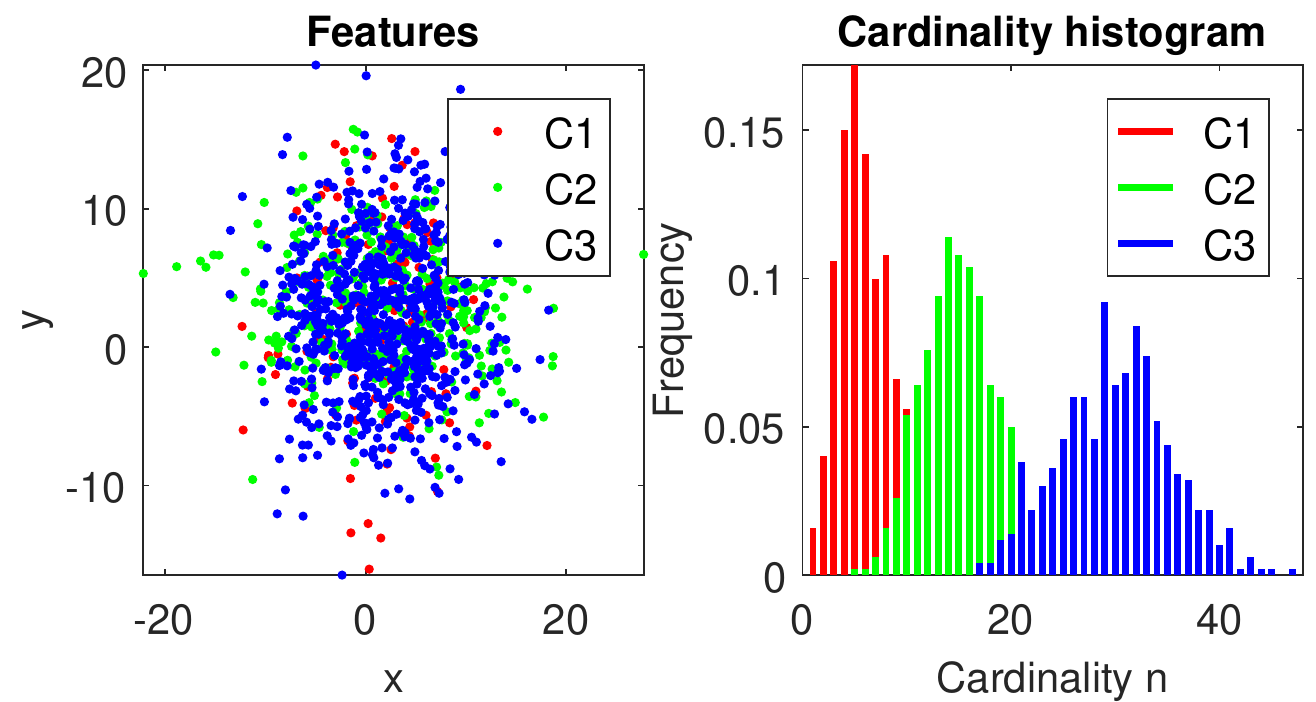}\vspace{-2mm}
\par\end{centering}
\caption{\label{fig:Sim_Classifi_feat_data_card_hist}Simulated data for the
experiment in section \ref{subsec:Classification-with-simulated}.
Left: Features of the data. Right: Cardinality histogram of the data.}

\vspace{-2mm}
\end{figure}

In this experiment, data are simulated from an underlying model consisting
of three clusters $C_{1}$, $C_{2}$ and $C_{3}$. A datum from $C_{i}$
is a finite set $X$ whose cardinality is Poisson distributed with
mean $\lambda_{i}$, and whose features are i.i.d. from a 2-D Gaussian
$\mathcal{N}(\cdot;\mu_{i},\Sigma_{i})$ where\vspace{-2mm}
\[
\begin{array}[t]{ccc}
\lambda_{1}=6,\,\, & \mu_{1}=[1,2]^{T}, & \Sigma_{1}=\mbox{diag}[20,40],\\
\lambda_{2}=15, & \mu_{2}=[2,3]^{T}, & \Sigma_{2}=\mbox{diag}[60,20],\\
\lambda_{3}=30, & \mu_{3}=[2,2]^{T}, & \Sigma_{3}=\mbox{diag}[30,30].
\end{array}
\]
Fig. \ref{fig:Sim_Classifi_feat_data_card_hist} plots the features
and cardinalities of data sampled from these models. 

Both the NB and RFS models are trained via ML estimation on a fully
observed training dataset set consisting of 900 data points (300 per
cluster). For NB, the data models are three 2-D Gaussians (one for
each cluster). For the RFS models, we use three Poisson RFSs with
2-D Gaussian feature distributions (which are in fact the same as
the Gaussians learned from the NB models). 

\begin{figure}[tbh]
\begin{centering}
\vspace{-3mm}
\subfloat[\label{fig:Sim_classifi_result} Simulated data]{\begin{centering}
\includegraphics[width=0.48\columnwidth]{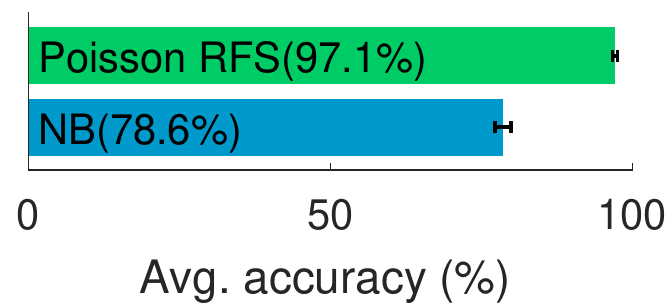}
\par\end{centering}

}~\subfloat[\label{fig:Texture_classifi_result} Texture data]{\begin{centering}
\includegraphics[width=0.48\columnwidth]{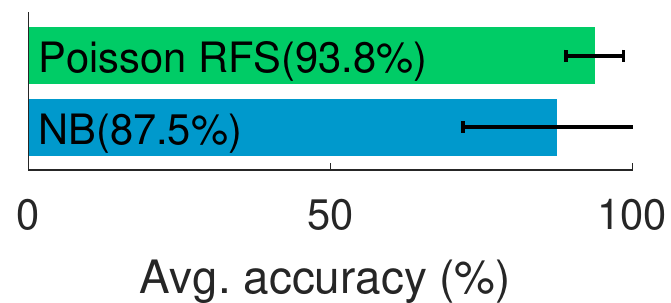}
\par\end{centering}

}
\par\end{centering}
\caption{Performance of classification by NB and Poisson RFS. Note that the
error-bars (on the peak of each column) are standard deviations of
accuracies, which show that Poisson RFS works more stably than NB.}
\vspace{-2mm}
\end{figure}

After training, we evaluate the learned models by classifying a test
set consisting of 1500 data points (500 per cluster, also simulated
from the described model). The evaluation are run 10 times with 10
different test sets, and the average accuracies\footnote{$\text{Accuracy}\triangleq\frac{\mbox{No. of correct classifications}}{\mbox{No. of observations in the test set}}$}
are shown in Fig. \ref{fig:Sim_classifi_result}. Observed that the
RFS model can exploit the cardinality information and delivers better
performance than NB. 

\subsubsection{Classification with real data}

\begin{figure}[tbh]
\begin{centering}
\vspace{-1mm}
\includegraphics[width=0.37\columnwidth]{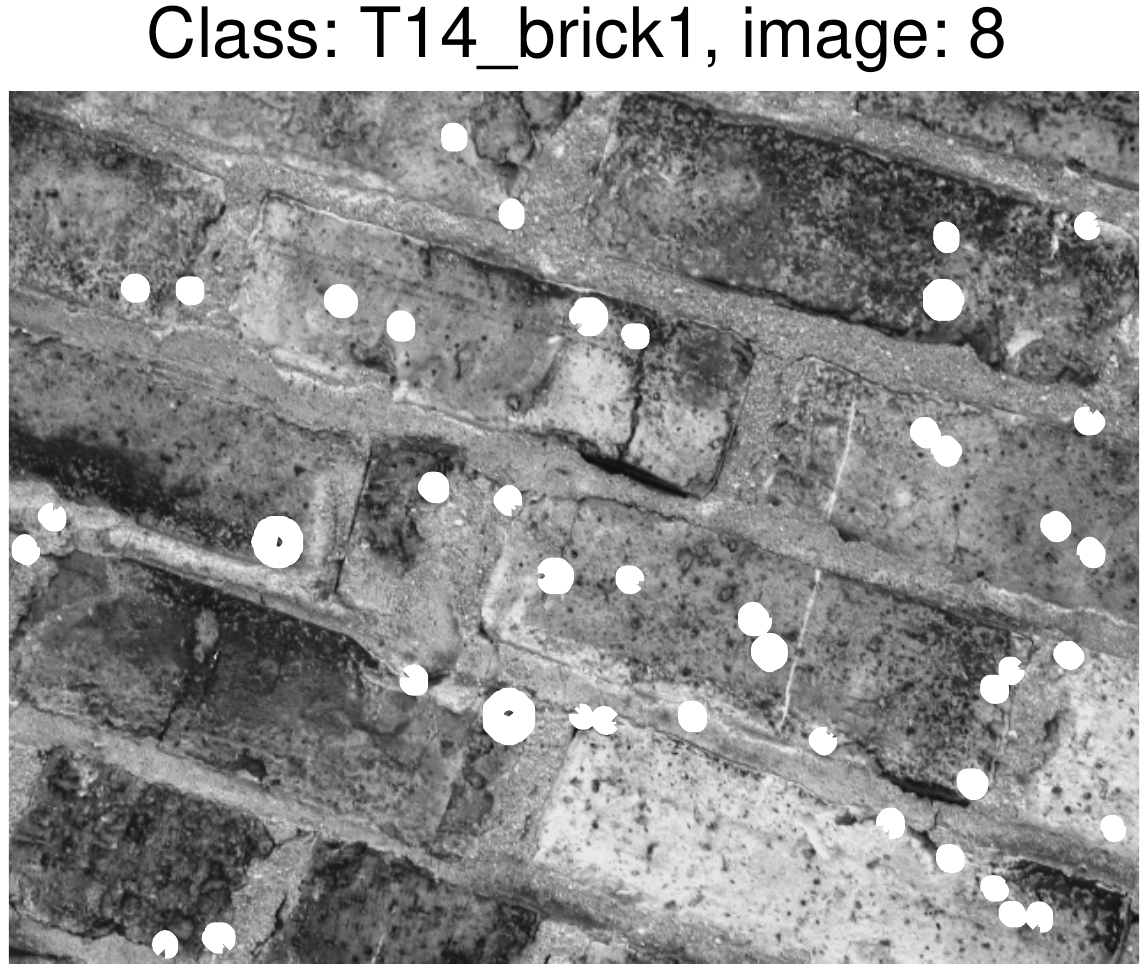}~\includegraphics[width=0.37\columnwidth]{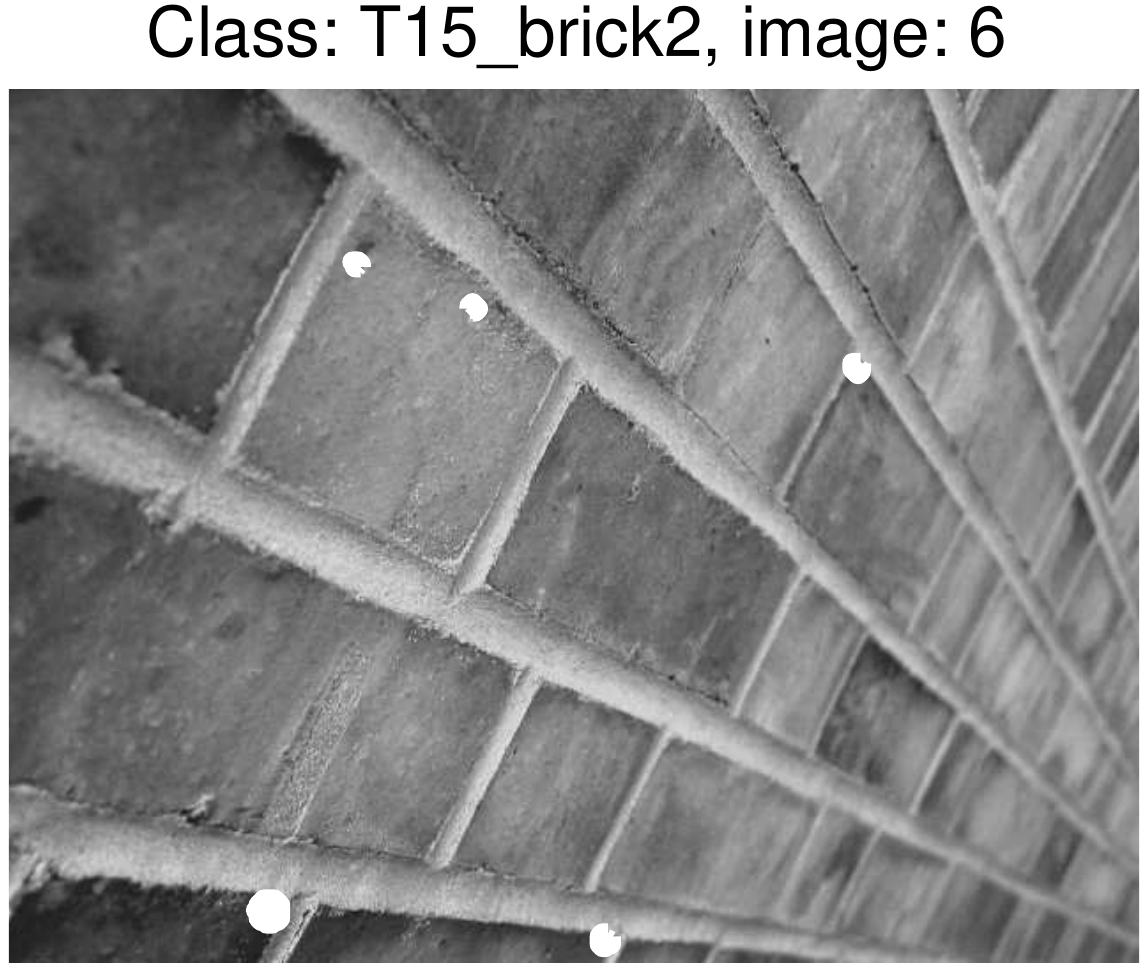}
\par\end{centering}
\begin{centering}
\vspace{-3mm}
\par\end{centering}
\caption{\label{fig:Texture_examples}Example images from 2 classes ``T15\_brick2''
and ``T14\_brick1'' in Texture dataset. Circles mark detected SIFT
keypoints.}
\vspace{-4mm}
\end{figure}

The second experiment involves classification of the two classes ``T14\_brick1''
and ``T15\_brick2'' from Texture images dataset \cite{textureDataset}.
Fig. \ref{fig:Texture_examples} show some example images from these
classes. 

\begin{figure}[tbh]
\begin{centering}
\vspace{-4mm}
\includegraphics[width=0.9\columnwidth]{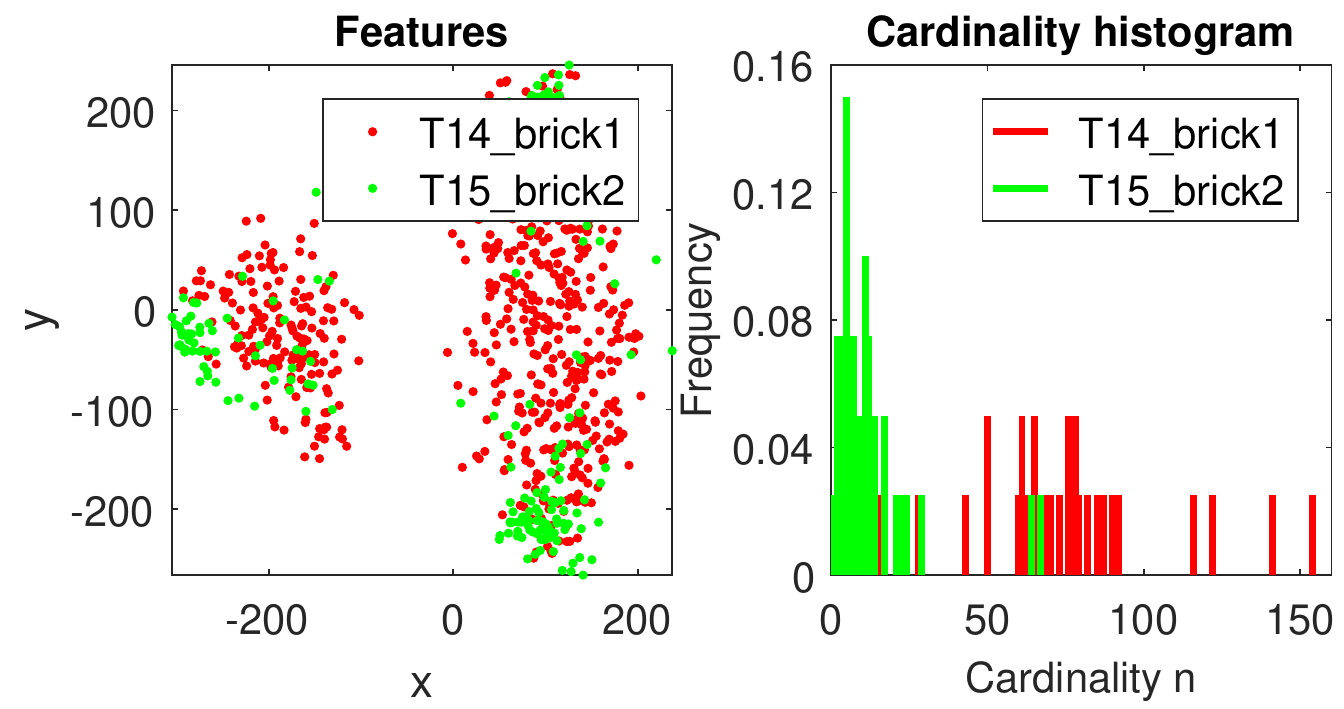}\vspace{-2mm}
\par\end{centering}
\caption{\label{fig:Texture_data}Extracted data from images of the Texture
dataset. Left: 2-D features (after applying PCA to the SIFT features).
Right: histogram of cardinalities of the extracted data. }

\vspace{-2mm}
\end{figure}

Features are extracted from each image by applying the SIFT algorithm\footnote{Using the VLFeat library \cite{vlfeatLib}.},
followed by Principal Component Analysis (PCA) to convert the 128-D
SIFT features into 2-D features. Thus each image is compressed into
a point pattern of 2-D features. Fig. \ref{fig:Texture_data} plots
the 2-D features of the images in the Texture images dataset.

The model parameters are learned by using MLE on a training dataset
containing 30 images per class. \textcolor{black}{For the NB model,
we use 2-D Gaussian distributions to model the data; and for the RFS
model, we use Poisson RFSs with 2-D Gaussian feature distributions.
 After training, the learned models are evaluated on a test set containing
all remaining images of the classes (10 images per class). The performance
is evaluated with 4-fold cross validation, and the average results
are shown in Fig. \ref{fig:Texture_classifi_result}. Observe that
the RFS model outperforms NB, since it can exploit cardinality information
of the data. }

\section{Ranking of Data\label{subsec:new_RFS_likeli}}

The previous section shows that RFS models avoid the unit inconsistency
and improve the classification performance by exploiting cardinality
information in point pattern data. However, using the RFS density
as the data ranking function for novelty detection does not result
in a good performance\footnote{We have done several experiments which are not shown here due to the
paper's length limitation.}. Due to the non-uniformity of the reference measure, the probability
densities for RFSs presented in the previous section do not provide
a consistent ranking of the data. In this section we propose a consistent
ranking function for the iid-cluster model. 

\subsection{Ranking Function \label{subsec:newRFSrank}}

Note that the probability density of an iid-cluster RFS (\ref{eq:iidRFSdensity})
is a product of $p_{f}^{X}$ and a term that depends only on the cardinality
$p_{c}(|X|)\,|X|!U{}^{|X|}$. Given the cardinality, $p_{f}^{X}$
is the density of the feature distribution relative to the Lebesgue
measure, and hence is the likelihood of $X$ given its cardinality.
However, $p_{c}(|X|)\,|X|!U{}^{|X|}$ is proportional to the Radon-Nikodym
derivative of the cardinality distribution relative to a Poisson distribution,
and thus does not indicate how likely the cardinality of $X$ is.
A ranking function $\ell$ that can reconcile this problem is:\vspace{-2mm}
\begin{equation}
\mathcal{\ell}(X)\propto p_{c}(|X|)\,[Cp_{f}]^{X}\label{eq:new_RFSlikeli_with_c}
\end{equation}
where $C$ is an unknown constant. 

To determine a suitable $C$, consider first the special case where
$p_{f}$ is uniform on a bounded state space $\mathcal{X}$. In such
case, all finite set subsets of $\mathcal{X}$ with the same cardinality
are equally likely, and a consistent ranking function should satisfy
$\ell(X)\propto p_{c}(m)$, given $|X|=m$. This condition can be
generalized to non-uniform $p_{f}$ by replacing $\ell(X)$ with its
expected value, given $|X|=m$
\begin{equation}
\mathbb{E}_{X}\left[\ell(X)\mid|X|=m\right]\propto p_{c}(m)\label{eq:conditional}
\end{equation}
Combining (\ref{eq:new_RFSlikeli_with_c}) with (\ref{eq:conditional})
and using the i.i.d. property of the features in iid-cluster models
yields:\vspace{-1mm}
\begin{equation}
\mathcal{\ell}(X)\propto p_{c}(|X|)\,\left[\frac{p_{f}}{||p_{f}||_{2}^{2}}\right]^{X}\label{eq:new_RFSlikeli_final}
\end{equation}
where $||p_{f}||_{2}^{2}=\int p_{f}^{2}(x)dx$ is the squared $L^{2}$-norm
of $p_{f}$, which has units of $U^{-1}$, and hence $\mathcal{\ell}(X)$
is unitless.

\subsection{Numerical experiments}

In this section we compare the performance of the proposed ranking
function (\ref{eq:new_RFSlikeli_final}) with the NB likelihood and
Poisson RFS likelihood in novelty detection. The threshold is set
at the $2^{nd}$ 10-quantile of the ranking function values\footnote{The performance depends on the (manually selected) threshold. Nonetheless,
the performance of the proposed ranking function would still be better
than the others since it can rank the data consistently as shown in
the boxplots of likelihood values for the experiment.} of the training dataset (consisting of only normal data). Observations
ranked below this threshold are classified as anomalies. 

\subsubsection{Novelty detection with simulated data}

In this experiment, normal data are samples, having cardinalities
between 40 and 60, drawn from a Poisson RFS with mean cardinality
48 and a 2-D Gaussian feature distribution\vspace{-1mm}
\[
\mathcal{N}\left(\cdot;\begin{bmatrix}0\\
0
\end{bmatrix},\begin{bmatrix}0.06 & 0.01\\
0.01 & 0.04
\end{bmatrix}\right).
\]
The same dataset containing 500 normal data points is used to train
the NB and Poisson RFS models with Gaussian feature distributions
via MLE (subsection \ref{subsec:Learn_RFS_params}).

\begin{figure}[tbh]
\begin{centering}
\vspace{-4mm}
\subfloat[\label{fig:Sim_anomaly_result} Simulated data]{\begin{centering}
\includegraphics[width=0.44\columnwidth]{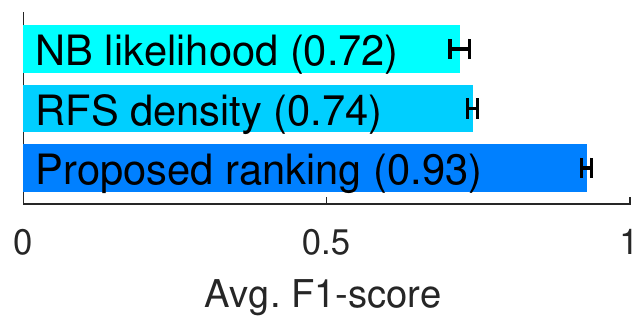}
\par\end{centering}

}\subfloat[\label{fig:Texture_anomaly_result} Texture data]{\begin{centering}
\includegraphics[width=0.44\columnwidth]{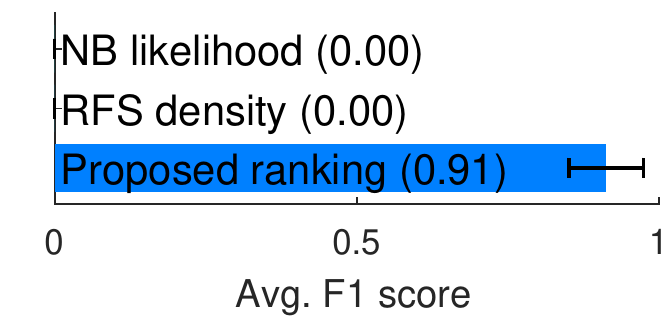}
\par\end{centering}

}\vspace{-2mm}
\par\end{centering}
\caption{Novelty detection results by 3 ranking functions: NB likelihood, Poisson
RFS likelihood, and proposed ranking function.}

\vspace{-1mm}
\end{figure}

Three types of anomalies are considered: \emph{low-cardinality anomaly}
(cardinality $\leq10$), \emph{high-cardinality anomaly} (cardinality
$\geq80$), and \emph{feature anomaly} which has the same cardinality
with normal data, but the Gaussian feature distribution now has a
mean of $[1,1]^{T}$. Novelty detection is performed on a test set
containing 200 normal observations and 300 anomalies (100 for each
type). The tests are run 10 times with 10 different randomly sampled
test sets. The averaged results shown in Fig. \ref{fig:Sim_anomaly_result}
indicated that the proposed ranking function yields superior performance.

\subsubsection{Novelty detection with real data}

This experiment uses the real dataset from the second experiment in
subsection \ref{subsec:experiments_RFS}. Normal data are taken from
the ``T14\_brick1'' class and anomalous test data are taken from
the ``T15\_brick2'' class. A 4-fold cross-validation is used: for
training we use 30 images from the normal class and for testing we
use the remaining images from normal class (10 images) and 10 images
from abnormal class (different at each time).

\begin{figure}[tbh]
\begin{centering}
\vspace{-2mm}
\includegraphics[width=0.86\columnwidth]{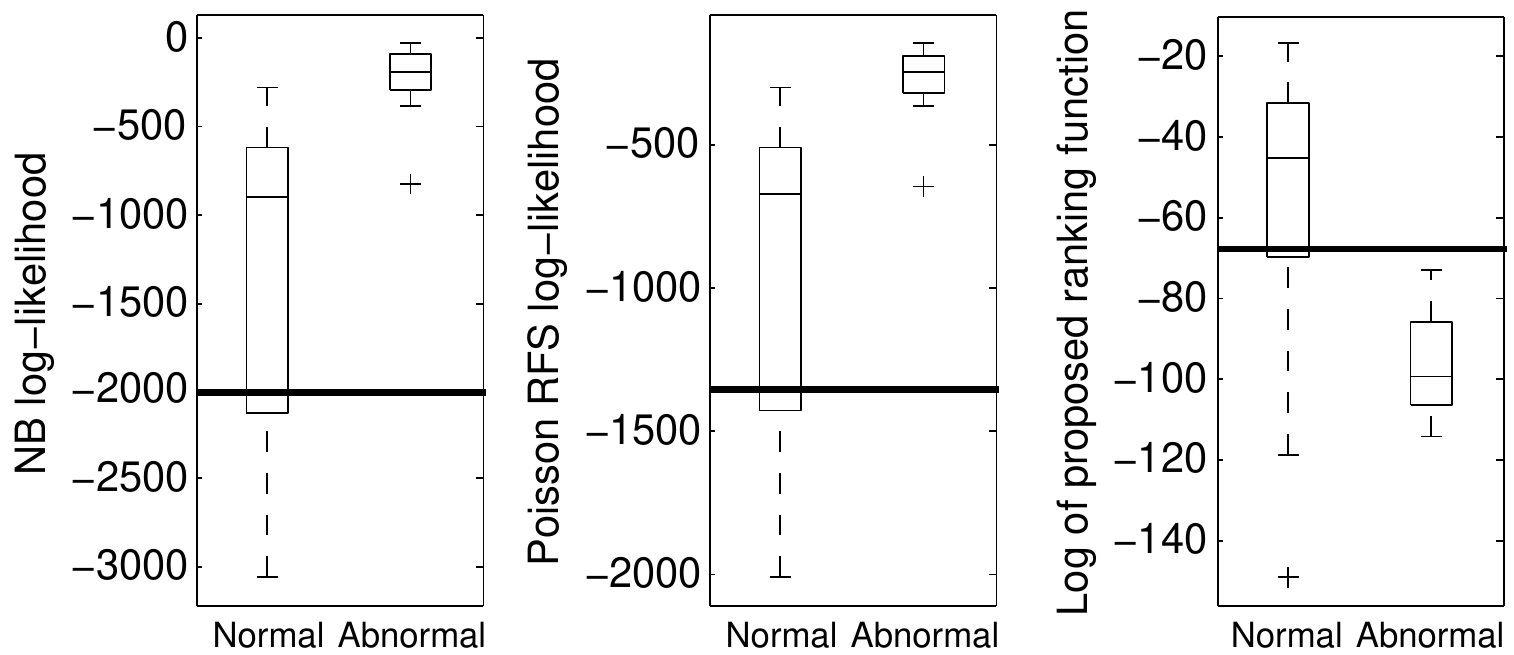}\vspace{-2mm}
\par\end{centering}
\caption{\label{fig:Texture_Boxplot_3models_and_result} Boxplots of likelihoods
computed by 3 models, namely NB likelihood, RFS density, and the proposed
ranking function, for a fold of Texture dataset. The solid line going
through each graph is the threshold (the $2^{nd}$ 10-quantile). }

\vspace{-2mm}
\end{figure}

As shown in Fig. \ref{fig:Texture_anomaly_result}, ranking the data
using the NB likelihood and Poisson RFS likelihood could not detect
any anomalies, while the proposed ranking function achieved an F1
score\footnote{$\mbox{\emph{F1 score}}\triangleq2\times\frac{\mbox{precision }\times\mbox{ recall}}{\mbox{precision }+\mbox{ recall}}$}
around 0.91. Moreover, Fig. \ref{fig:Texture_Boxplot_3models_and_result}
verified that only the proposed ranking function provides a consistent
ranking, while the NB and Poisson RFS likelihoods even rank anomalies
higher than normal data. 

\section{Conclusions}

In this paper, we have introduced statistical models for point pattern
data using Random Finite Set theory. Such models provide the means
for developing model-based classification and novelty detection for
point pattern data. In particular we proposed a maximum likelihood
method for learning the parameters of an iid-cluster RFS\textendash the
analogue of the naïve Bayes model for point patterns. For novelty
detection, we proposed novel ranking functions based on RFS models,
which substantially improve performance. Our results also contribute
to the Bag-Space paradigm in multiple instance learning where statistical
models are not available. 

\bibliographystyle{IEEEtran}
\bibliography{Refs/ref1}

\end{document}